\documentclass[letterpaper, 10 pt, conference]{ieeeconf}
\IEEEoverridecommandlockouts

\usepackage{float}
\usepackage{tabularx}
\newcolumntype{Y}{>{\raggedright\arraybackslash}X}

\usepackage{xspace}
\usepackage{xcolor}
\usepackage{amsfonts}
\usepackage{graphics} %
\graphicspath{{figures/}}  %
\usepackage{epsfig} %
\usepackage{mathptmx} %
\usepackage{times} %
\usepackage{amsmath} %
\usepackage{amssymb}  %
\usepackage{ulem}
\usepackage{siunitx}
\usepackage{verbatim}
\usepackage{booktabs}
\usepackage{multirow}
\usepackage[space]{cite}
\usepackage[table]{xcolor} %
\definecolor{lightgreen}{RGB}{210,255,210}
\usepackage[colorlinks=true, linkcolor=blue, urlcolor=blue, citecolor=blue]{hyperref}
\usepackage{makecell}
\usepackage{lipsum}

\usepackage[font=footnotesize]{caption}

\makeatletter
\let\origthebibliography\thebibliography
\def\thebibliography#1{%
  \origthebibliography{#1}%
  \scriptsize %
}
\makeatother

\newcommand{\method}{ExStereo\xspace}

\definecolor{figgray}{rgb}{0.2, 0.2, 0.2}
\definecolor{figgreen}{rgb}{0.280, 0.594, 0.296}
\definecolor{figblue}{rgb}{0.082, 0.396, 0.753}
\definecolor{figlightred}{rgb}{1.0, 0.149, 0.0}
\definecolor{figdarkblue}{rgb}{0.188, 0.392, 0.729}

\title{\LARGE \bf
\method: Lifting 2D Vision-Language-Action Models to 3D\\with Explicit Stereo Representations
}

\author{
    I-Chun~Arthur Liu\textsuperscript{*}
    \quad
    Jason Chen\textsuperscript{*}
    \thanks{\textsuperscript{*}Equal contribution.}
    \quad
    Gaurav~S.~Sukhatme
    \quad
    Daniel Seita\\
    Viterbi School of Engineering, University of Southern California
}

\begin{document}
\maketitle

\begin{abstract}
Three-dimensional perception is critical for robotic manipulation, particularly for high-precision tasks, as recovering metric depth and precise 3D object positions from monocular RGB observations is inherently ill-posed. However, many Vision-Language-Action (VLA) models rely solely on RGB observations for perception. Leveraging recent advances in foundation models for stereo matching, we introduce \method, a stereo module that augments pre-trained 2D VLAs with 3D perception. \method reconstructs scene geometry from stereo image pairs and renders multi-view observations as an explicit stereo representation for stereo feature extraction. The action tokens from the action expert selectively attend to the resulting stereo tokens through our proposed action-stereo cross-attention mechanism, enabling the policy to generate robot actions conditioned on 3D scene information. To learn robust 3D representations, we introduce a mid-training stage before task-specific post-training, using a self-supervised learning objective on large-scale stereo data. We validate our approach by fine-tuning two publicly available VLAs, $\pi_{0.5}$ and SmolVLA, and evaluate them in simulation and on a real-world bimanual PiPER platform. Across both settings, VLAs fine-tuned with \method consistently outperform baselines, demonstrating the effectiveness of stereo perception for robotic manipulation. Our anonymous project website is available at \href{https://exstereo-vla.github.io/ExStereo/}{https://exstereo-vla.github.io/ExStereo/}.

\end{abstract}

\section{Introduction}

Vision-Language-Action (VLA) models have recently emerged as a powerful paradigm for robotic manipulation~\cite{kim24openvla,intelligence2025pi05,liu2024rdt}, demonstrating impressive visuomotor capabilities. By building on large-scale pretrained Vision-Language Models (VLMs)~\cite{radford2021clip,dosovitskiy2020vit}, these policies learn from semantic priors, acquire diverse skills from demonstrations, and generalize across objects and instructions. However, most VLA models operate on monocular RGB observations processed through pre-trained 2D vision encoders to extract features for downstream action generation. Operating on 2D visual tokens, these policies lack precise depth and geometric information. While effective in controlled settings, real-world interactions often require accurate spatial relationships and object locations, as well as robustness to occlusions and visual variations.

Recent work has explored injecting 3D structure into VLAs~\cite{li2025bridgevla,li2025pointvlainjecting3dworld,sun2025geovlaempowering3drepresentations,zhen20243dvla}, achieving strong performance on 3D manipulation tasks. However, these methods depend on RGB-D sensors, which require precise calibration and are sensitive to sensor noise. Meanwhile, StereoVLA~\cite{deng2026stereovlaenhancingvisionlanguageactionmodels} and StereoPolicy~\cite{han2026stereopolicyimprovingroboticmanipulation} leverage stereo image pairs to obtain implicit stereo features for policy learning. StereoVLA requires training its VLA from scratch, while StereoPolicy provides only modest improvements to pre-trained VLAs. In this work, we show how effectively leveraging explicit stereo representations can equip 2D VLAs with metric 3D perception and substantially improve their performance.

\begin{figure}[t]
    \centering
    \includegraphics[width=\columnwidth]{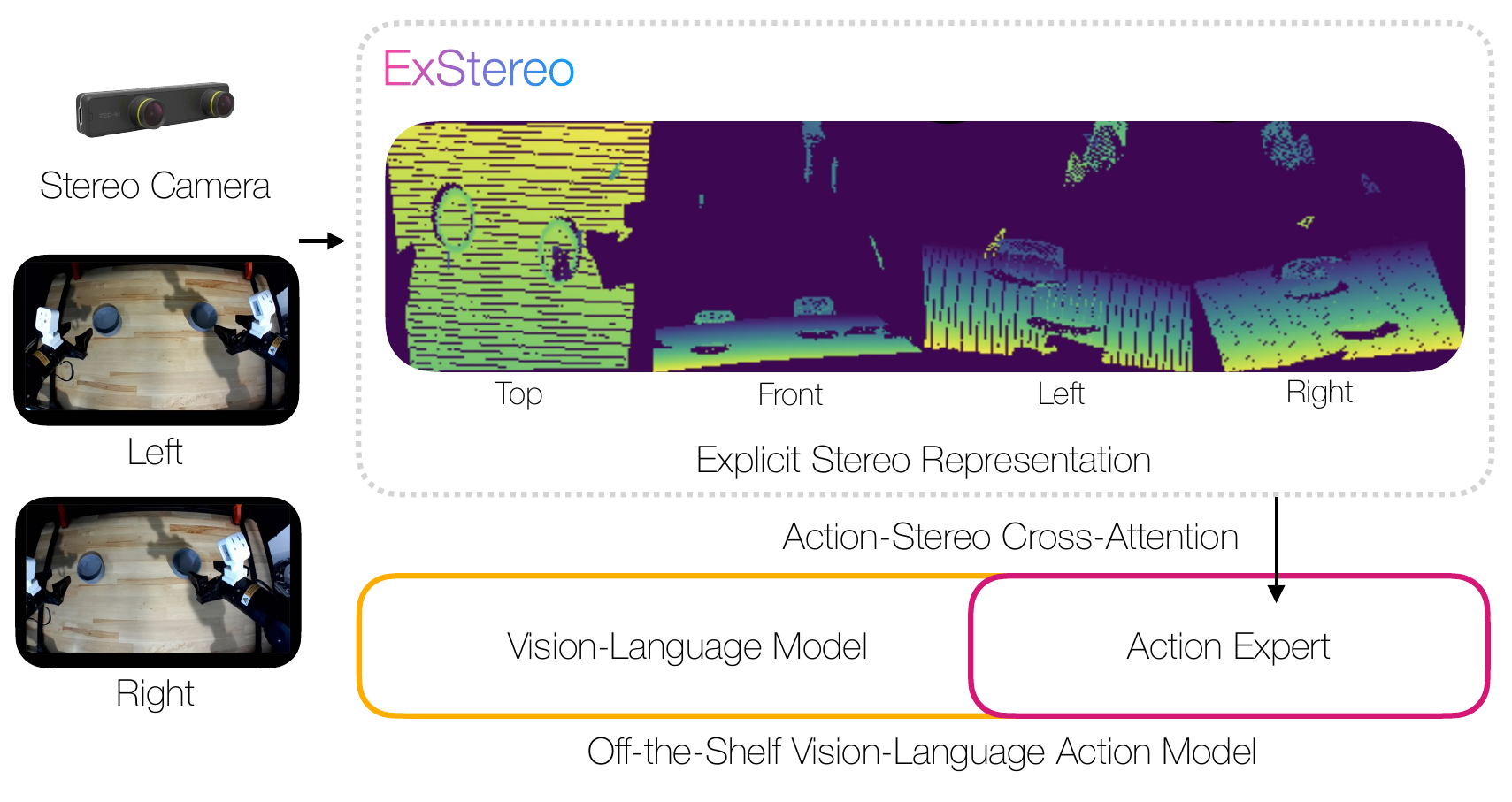}
    \caption{\method is a stereo module that enables 3D spatial perception in pre-trained, off-the-shelf Vision-Language-Action models. Given a stereo image pair, it constructs four orthogonal views of the scene with metric 3D coordinates. This explicit 3D representation allows the models to selectively retrieve visual features relevant to the action tokens through our proposed action-stereo cross-attention mechanism, improving robotic manipulation performance.}
    \label{fig:hero_figure}
\end{figure}

\begin{figure*}[t]
    \centering
\includegraphics[width=1.0\textwidth]{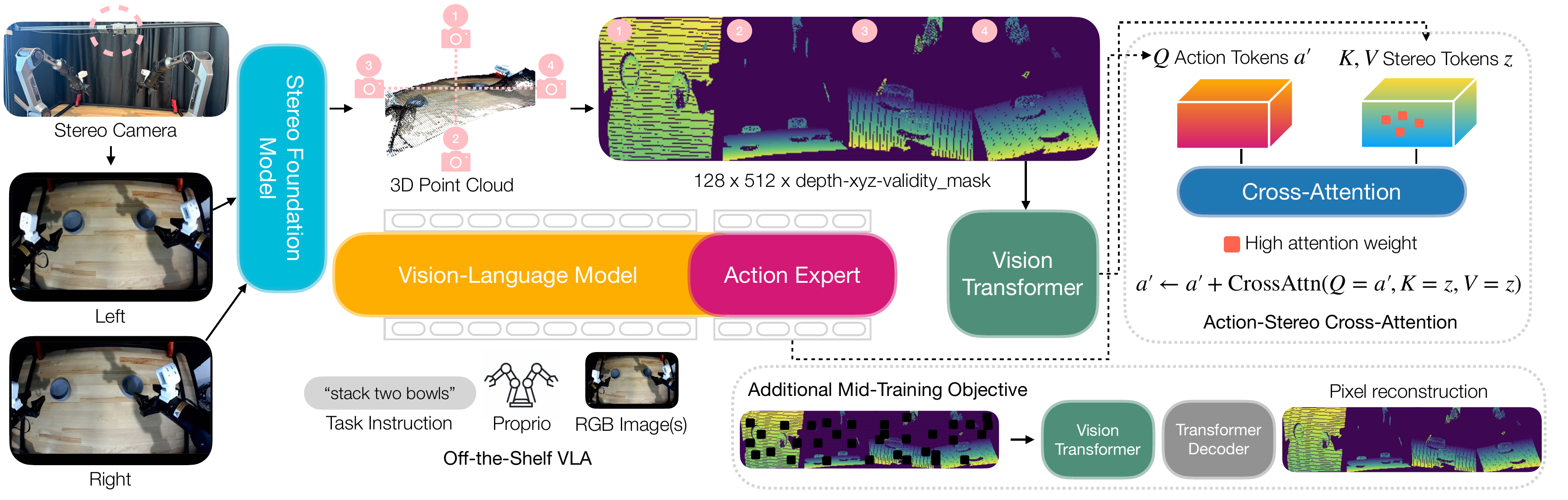}
    \caption{\textbf{Overview of \method.} 
    \method enhances pre-trained VLA models by leveraging a stereo camera and a pre-trained, frozen stereo foundation model~\cite{wen2026fastfoundationstereo}. Given a stereo image pair, the stereo model predicts a disparity map, which is converted into a metric depth map and 3D point cloud using the camera intrinsics and stereo baseline. We render the point cloud from four orthogonal viewpoints (Top, Front, Left, and Right) using z-buffering. Each view is rasterized at a resolution of $128 \times 128$ and contains depth, 3D coordinates, and a validity mask. The four views are horizontally tiled into a multi-view observation of size $128 \times 512 \times 5$, then processed by a Vision Transformer (ViT)~\cite{dosovitskiy2020vit} to extract stereo tokens $z$.
    We leverage $z$ for action generation through our proposed action-stereo cross-attention, where action tokens $a'$ query $z$ to selectively retrieve relevant 3D visual features while leaving the VLA's prefix sequence and VLM weights unchanged. To learn robust 3D representations, we introduce mid-training with masked autoencoding (MAE), where the VLA reconstructs randomly masked multi-view patches from a large-scale stereo dataset while learning action-relevant features from incomplete observations. During VLA post-training, the ViT and action-stereo cross-attention block are frozen to prevent overfitting to task-specific demonstrations, while the MAE decoder is discarded.
    }
    \label{fig:method}
\end{figure*}

We propose lifting 2D Vision-Language-Action models to 3D with \underline{\textbf{Ex}}plicit \underline{\textbf{Stereo}} Representations (ExStereo), a stereo module that equips pre-trained, off-the-shelf VLAs with 3D spatial perception. It takes a stereo image pair as input and employs a pre-trained stereo foundation model~\cite{wen2026fastfoundationstereo} to obtain a disparity map. Using the camera intrinsics and stereo baseline, the disparity map is transformed into a metric depth map and 3D point cloud. The point cloud is rendered from four orthogonal viewpoints (Top, Front, Left, and Right) using z-buffering. The four views are then tiled horizontally to form a multi-view observation, which is fed into a Vision Transformer (ViT)~\cite{dosovitskiy2020vit} to obtain stereo tokens. The stereo tokens are utilized for action generation through the proposed action-stereo cross-attention mechanism, where the action tokens serve as queries and the stereo tokens serve as keys and values. This allows the action tokens to selectively attend to relevant 3D visual features without modifying the large prefix token sequence or the weights of VLMs, enabling pre-trained robot policies to efficiently integrate the stereo module during post-training. See~\autoref{fig:hero_figure} for an overview.

To learn robust 3D representations, we introduce a \emph{mid-training} stage with a masked autoencoding objective prior to post-training. During mid-training, patches of the multi-view observation are randomly masked, and the VLA is trained on a large-scale stereo dataset to reconstruct the missing patches while learning to extract relevant visual features for action generation from incomplete observations. During post-training, the stereo module is frozen to prevent overfitting to task-specific demonstrations. 
We evaluate ExStereo on ten simulated RoboTwin~\cite{chen2025robotwin} tasks and three real-world tasks using a bimanual AgileX PiPER platform. ExStereo improves the average simulation success rate of $\pi_{0.5}$ from 51.2\% to 76.4\% and its real-world success rate from 16/60 to 50/60 trials, while also outperforming point-cloud and implicit-stereo baselines.

The contributions of this paper are:
\begin{itemize}
    \item We propose \method, a novel stereo module that lifts any 2D VLA to 3D for fast adaptation without requiring additional pre-training of the base policy.
    \item We demonstrate that explicit stereo representations, when used appropriately, yield stronger downstream manipulation performance than implicit representations.
    \item Extensive simulation and real-world experiments show that VLAs equipped with ExStereo greatly improve upon the performance of their base VLAs while outperforming other strong baselines.
\end{itemize}

\section{Related Work}

\subsection{Vision-Language-Action Models}
\label{sec:related_VLA}

Vision-Language-Action (VLA) models, also commonly referred to as robotic foundation models, fine-tune pretrained VLMs to output robot actions, such as end-effector pose deltas or joint targets, by training on large-scale teleoperation data~\cite{open_x_embodiment_rt_x_2023}. Representative systems include RT-2, OpenVLA~\cite{kim24openvla}, SmolVLA~\cite{shukor2025smolvlavisionlanguageactionmodelaffordable}, and the $\pi$ family~\cite{black2024pi0visionlanguageactionflowmodel, intelligence2025pi05,intelligence2025pi06vlalearnsexperience,intelligence2026pi07}. We refer the reader to~\cite{FMs_robotics_2023_survey_1,FMs_robotics_2023_survey_2} for surveys. 
Early VLAs discretized actions into tokens and predicted them autoregressively~\cite{rt22023arxiv, kim24openvla}, whereas recent policies attach a dedicated action expert trained with diffusion or flow matching to produce smooth, high-frequency action chunks~\cite{black2024pi0visionlanguageactionflowmodel, intelligence2025pi05,shukor2025smolvlavisionlanguageactionmodelaffordable}. 

Camera configurations have broadened in parallel, from a single third-person view to wrist-mounted cameras and to multi-camera rigs that observe the workspace from several sides~\cite{intelligence2025pi05}. Across these variants, each view is still processed by a monocular 2D encoder so metric scale and inter-object spatial relations must be inferred from appearance. Prior work has addressed these limitations by incorporating depth sensors~\cite{liu2026clamp,jia2024lift3dfoundationpolicylifting,wei2024droma}, such as RGB-D cameras. However, these sensors often degrade on transparent~\cite{wei2024droma} and reflective surfaces and are sensitive to sensor noise, calibration errors, and point cloud artifacts, making high-quality 3D reconstruction challenging. Beyond these hardware limitations, visual representations remain a key bottleneck. Existing 2D vision backbones~\cite{radford2021clip,Tschannen2025SigLIP2M,simeoni2025dinov3} generalize well across diverse visual domains but lack explicit 3D structure. In contrast, 3D vision backbones directly capture geometry from point clouds~\cite{wu2025sonata,wu2023masked,wu2024ptv3,PointContrast2020}, but tend to underperform their 2D counterparts due to the scarcity of large-scale 3D training data.

These limitations motivate a 3D perception approach that preserves the scalability and representational strengths of 2D vision while providing explicit geometric information. Stereo setups are common on robot platforms and can recover such geometry from binocular disparity, yet the second image is rarely exploited by the policy. ExStereo builds on this observation and equips pre-trained VLAs with an explicit 3D representation computed from a stereo pair.

\subsection{Stereo Vision in Computer Vision}
\label{sec:related_stereo}
Stereo matching recovers metric depth by estimating disparity between a rectified left--right
image pair without active depth sensors. Previous classical methods build a cost volume and refine
disparity with 3D convolutions or recurrent updates~\cite{chang2018pyramidstereomatchingnetwork,lipson2021raft,xu2025igev}.
More recently, the field has moved toward stereo foundation models that generalize zero-shot
across domains by transferring priors from monocular depth models. FoundationStereo~\cite{wen2025stereo}
establishes this direction through large-scale pretraining, and later work improves robustness
on transparent, reflective, and textureless surfaces where depth sensors
fail~\cite{Bartolomei_2025_CVPR,guo2024stereo}. Fast-FoundationStereo~\cite{wen2026fastfoundationstereo}
distills these models for near real-time inference. We leverage it in \method to obtain an explicit 3D representation.

\subsection{Stereo Vision in Robotics}
Stereo cameras recover depth from binocular disparity, making them more reliable for transparent, specular, and textureless objects, where conventional depth sensors often struggle. Prior work has applied stereo vision to challenging materials such as glass~\cite{wei2024droma,bai2024cleardepth}, as well as to stereo-based grasping and tracking~\cite{yu2025pogs}. More recently, StereoPolicy~\cite{han2026stereopolicyimprovingroboticmanipulation} and StereoVLA~\cite{deng2026stereovlaenhancingvisionlanguageactionmodels} extract features from stereo modules, which we characterize as \emph{implicit} representations (see Section \ref{sec:multiviewobservationrendering} for details), and combine them with language or RGB image features as inputs to a VLM for policy learning. In contrast, we use a stereo module to construct an explicit 3D representation of the scene and feed the corresponding tokens directly into the action expert for policy learning.

\section{Problem Statement}
Given a pre-trained VLA model $\pi_\theta$, parameterized by $\theta$, that maps an image observation $I_t$, proprioceptive state $s_t$, and language instruction $\ell$ to an action chunk $a_{t:t+H} = \pi_\theta(I_t, s_t, \ell)$, where $H$ denotes the action chunk size, our goal is to augment $\pi_\theta$ with a stereo module $\psi$ that perceives the scene through a calibrated stereo camera. The stereo camera provides a left image $I_t^L$ and a right image $I_t^R$, rectified using camera intrinsics $K$ and baseline $B$. The augmented VLA generates an action as $a_{t:t+H} = \pi^{\psi}_{\theta}\!\left(I^{L}_{t}, I^{R}_{t}, s_{t}, \ell\right)$, 
where the stereo module $\psi$ comprises the frozen stereo foundation model, the geometric reconstruction and rendering pipeline, the multi-view encoder $\rho$, and the action-stereo cross-attention block introduced in the action expert. 
Each dataset $\mathcal{D}$ consists of expert demonstrations of the form $E = \left(I_1^L, I_1^R, a_1, \ldots, I_T^L, I_T^R, a_T\right)$, where $T$ denotes the number of time steps in each demonstration. We use the term VLA ``fine-tuning'' to refer to both mid-training and post-training, encompassing any additional training performed on the provided pre-trained VLA.

\section{Method: \method}

\method is the stereo module $\psi$ that takes a stereo image pair as input, generates stereo tokens, and fuses them with action tokens from the VLA for 3D-conditioned action generation. It first constructs a 3D scene point cloud for multi-view observation rendering (Section~\ref{sec:multiviewobservationrendering}). The rendered multi-view observations are then fed into a point cloud encoder $\rho$ to obtain multi-view tokens, which are cross-attended to queries from the action tokens. This allows each action timestep to selectively attend to any of the multi-view tokens (Section~\ref{sec:stereoperceptionmodulearchitecture}). In Section~\ref{sec:pretrainingandfinetuning}, we describe the two-stage training process. 
See~\autoref{fig:method} for an overview.

\subsection{Multi-View Observation Rendering}
\label{sec:multiviewobservationrendering}
For a rectified stereo pair image $(I^{L}_{t}, I^{R}_{t})$, we input it into a pre-trained, frozen Fast-FoundationStereo\cite{wen2026fastfoundationstereo} $\mathcal{F}$, a foundation model for stereo depth estimation, to obtain a dense disparity map
$d_{t} = \mathcal{F}(I^{L}_{t}, I^{R}_{t})$. We use the disparity map, horizontal focal length $f_{x}$ from $K$, and $B$ to compute metric depth $Z_{t} = \frac{f_x B}{d_{t}}$. We then back-project the depth map using $K$ to obtain a metric 3D point cloud in the left camera coordinate frame.

To ensure consistent multi-view observations, we align the point cloud to a shared world coordinate frame: camera coordinates are remapped to a conventional world frame ($X$ right, $Y$ forward, $Z$ up). The dominant plane, corresponding to the table surface, is detected using RANSAC~\cite{fischler_ransac} and rotated to be level such that its normal is aligned with the $+Z$ axis. The table surface is then shifted to $z=0$, and the point cloud is horizontally re-centered by mean-centering the $X$ and $Y$ coordinates. We place four ``virtual'' cameras at orthogonal locations around the origin and render the point cloud into four orthographic views (Top, Front, Left, and Right) using a z-buffer that retains the nearest surface at each pixel. Each view is rasterized at a resolution of $128 \times 128$ and contains depth, 3D coordinates, and a validity mask. Note that the depth values are expressed in each virtual view’s coordinate frame, whereas the 3D coordinates are expressed in the world frame. The four views are concatenated horizontally to form the multi-view observation $O_t \in \mathbb{R}^{128 \times 512 \times 5}$.

We define the explicit reconstruction of the point cloud and its multi-view rendering as an \emph{explicit} stereo representation, in contrast to \emph{implicit} stereo representations, which directly process stereo image pairs through a stereo module without explicitly reconstructing the underlying 3D geometry. This explicit representation enables the policy to perceive and reason about the environment in metric 3D space.

\subsection{Multi-View Encoding and Action–Stereo Cross-Attention}
\label{sec:stereoperceptionmodulearchitecture}

We encode the multi-view observation $O_t$ using a Vision Transformer (ViT) $\rho$~\cite{dosovitskiy2020vit}. The input is patchified using a strided convolution with a patch size of $16 \times 16$, yielding a total of $256$ stereo tokens. Learned positional embeddings are added to the tokens, and neither a class token nor pooling is used. The encoder outputs the unpooled stereo tokens:
\begin{equation}
z_t = \rho(O_t) \in \mathbb{R}^{256 \times d},
\end{equation}
where $d$ denotes the hidden dimension of the ViT.

To effectively integrate the stereo module into a pre-trained VLA model, we propose fusing the action tokens with the stereo tokens using a single multi-head cross-attention block. We refer to this mechanism as action-stereo cross-attention. Given action tokens $a_t' \in \mathbb{R}^{H \times d}$ generated from the VLA and stereo tokens $z_t$, we construct queries from the action tokens and keys and values from the stereo tokens:
\begin{equation}
Q = a_t' W^Q, \qquad
K = z_t W^K, \qquad
V = z_t W^V.
\end{equation}
For each attention head $i$, we compute scaled dot-product attention as
\begin{equation}
A_i =
\mathrm{softmax}\!\left(
\frac{Q_i K_i^\top}{\sqrt{d_h}}
\right)V_i,
\qquad i = 1,\dots,h,
\label{eq:lf-scores}
\end{equation}
where $h$ is the number of attention heads and $d_h$ is the dimension of each head. The outputs from all attention heads are concatenated and projected through a learned output matrix $W^O$ to obtain the stereo context for each action token. The resulting context is then added residually to the action tokens:
\begin{equation}
\begin{aligned}
a_t' \leftarrow {} & a_t' +\\
&\mathrm{Concat}_{i=1}^{h} \Bigg[
\mathrm{softmax}\!\left(
\frac{(a_t' W^Q)_i (z_t W^K)_i^\top}
{\sqrt{d_h}}
\right)
(z_t W^V)_i
\Bigg] W^O.
\end{aligned}
\label{eq:lf-compact}
\end{equation}
As the cross-attention block operates entirely on the action tokens and never modifies the prefix tokens or weights of the VLM, this stereo-conditioning mechanism is agnostic to the choice of VLM backbone and can be integrated into any VLA architecture that maintains an action-generation module.

\subsection{Mid-Training and Post-Training}
\label{sec:pretrainingandfinetuning}
For VLA fine-tuning, we train $\pi_{\theta}$ together with $\psi$ in two stages: \emph{mid-training} and \emph{post-training}. Mid-training is an intermediate stage between pre-training a VLA on broad data and post-training on task-specific demonstrations. As shown empirically in our experiments, mid-training is particularly important for methods that introduce a new, unseen module after pre-training, as it allows the module to gradually adapt during policy learning. In this work, we mid-train $\pi_{\theta}$ together with $\psi$ on a large-scale stereo dataset to learn generalizable 3D representations for bimanual manipulation tasks.

During mid-training, we leverage StereoEngine~\cite{chen2027stereoengine} to generate a large-scale stereo dataset for real-world experiments. StereoEngine is a recent method that uses video diffusion to either lift existing monocular demonstrations into stereo pairs or directly generate both stereo views from simulation. In addition to the original VLA training objective (e.g., the flow-matching action loss), we introduce a masked autoencoding (MAE)~\cite{He2021MaskedAA} objective on the multi-view observations $O_t$.

Specifically, at each training step, we uniformly sample and mask a subset of the $16 \times 16$ patches in $O_t$, producing a masked multi-view observation $\tilde{O}_t$. The embedding of each masked patch is replaced with a learned mask token, while the total number of tokens remains unchanged. The resulting output of $\rho$, denoted by $z'_t$, therefore contains both visible and masked tokens. We provide $z'_t$ to both the cross-attention block for action generation and a Transformer decoder $G_{\mathrm{mae}}$ that reconstructs the depth, 3D coordinates, and validity mask of the masked patches.

This objective encourages the policy to learn robust 3D representations from incomplete observations, which can improve robustness to occlusions, distractors, and sensor noise. Masked reconstruction has also been shown to produce representations that transfer effectively to downstream vision tasks~\cite{He2021MaskedAA}; here, we leverage this property to learn 3D representations useful for action generation. The MAE objective is defined as
\begin{equation}
\mathcal{L}_{\mathrm{MAE}}
=
\mathbb{E}_{O_t \sim \mathcal{D}_{\mathrm{mid}}}
\left[
\left\|
m \odot
\left(
G_{\mathrm{mae}}(z'_t) - O_t
\right)
\right\|_2^2
\right],
\end{equation}
where $m$ selects the masked patches. The overall mid-training objective is
\begin{equation}
\mathcal{L}_{\mathrm{Total}}
=
\mathcal{L}_{\mathrm{VLA}}
+
\lambda \mathcal{L}_{\mathrm{MAE}},
\end{equation}
where $\lambda$ is a small fixed weighting coefficient.

\begin{table*}
  \setlength\tabcolsep{5pt}
  \centering
    \begin{tabular}{l ccc c >{\columncolor{lightgreen}}c >{\columncolor{lightgreen}}c c >{\columncolor{lightgreen}}c}
    \toprule
        & \multicolumn{3}{c}{Baselines} & \multicolumn{3}{c}{$\pi_{0.5}$} & \multicolumn{2}{c}{SmolVLA} \\
    \cmidrule(lr){2-4} \cmidrule(lr){5-7} \cmidrule(lr){8-9}
        Task & ACT & DP3 & StereoVLA & Base & \makecell{\method\\(w/o MT)} & \method & Base & \method \\
    \midrule
    Lift Pot            & 2.7  & 5.3  & 0.0  & 58.7 & \textbf{66.7} & 61.3          & 9.3  & \textbf{12.0} \\
    Adjust Bottle       & 10.7 & 2.7  & 53.3  & 76.0 & 89.3          & \textbf{96.0} & 58.7 & \textbf{96.0} \\
    Grab Roller         & 42.7 & 5.3  & 17.3 & 80.0 & 89.3          & \textbf{100.0}& \textbf{57.3} & 48.0 \\
    Open Laptop         & 0.0  & 5.3  & 2.7  & 61.3 & 72.0          & \textbf{92.0} & 2.7  & \textbf{44.0} \\
    Beat Block Hammer   & 0.0  & 4.0  & 0.0  & 25.3 & 45.3          & \textbf{70.7} & 1.3  & \textbf{24.0} \\
    Click Alarmclock    & 2.7  & 10.7 & 13.3  & 21.3 & 24.0          & \textbf{54.7} & 14.7 & \textbf{22.7} \\
    Dump Bin Bigbin     & 2.7  & 60.0 & 30.7  & 58.7 & 66.7          & \textbf{77.3} & 26.7 & \textbf{49.3} \\
    Handover Mic        & 0.0  & 28.0 & 6.7  & 26.7 & 49.3          & \textbf{77.3} & 0.0  & \textbf{65.3} \\
    Press Stapler       & 10.7 & 14.7 & 10.7  & 76.0 & \textbf{80.0} & 78.7          & 14.7 & \textbf{53.3} \\
    Stamp Seal          & 0.0  & 0.0  & 0.0  & 28.0 & 45.3          & \textbf{56.0} & 1.3  & \textbf{5.3}  \\
    \midrule
    Average             & 7.2  & 13.6 & 13.5  & 51.2 & 62.8          & \textbf{76.4} & 18.7 & \textbf{42.0} \\
    \bottomrule
    \end{tabular}
    \caption{
        \textbf{Simulation Results.} Per-task success rates (\%) on ten bimanual RoboTwin tasks, averaged over three random seeds. We fine-tune two pre-trained VLAs, $\pi_{0.5}$ and SmolVLA, with \method (highlighted in \textcolor{figgreen}{green}) and compare them against their fine-tuned base policies. ACT (RGB), DP3 (point cloud), and StereoVLA serve as additional baselines. ``w/o MT'' denotes \method without the mid-training stage. For each base VLA, the best result for each task is shown in \textbf{bold}. See Section~\ref{sec:sim_exps} for more details. 
    }
  \vspace*{-10pt}
  \label{tab:sim-results}
\end{table*}

In post-training, we initialize the policy with the weights from the mid-training stage and keep the stereo module $\psi$ frozen. The policy is then trained on a limited amount of task demonstrations $\mathcal{D}_{\text{post}}$ using its standard Low-Rank Adaptation (LoRA) recipe~\cite{hu2022loralowrankadaptationlarge}. Note that the MAE objective is not used during this stage, and the MAE decoder is discarded. This configuration preserves the general 3D representations learned at scale while preventing the stereo module from overfitting to the limited task-specific demonstrations.

\begin{figure}[t]
    \centering
    \includegraphics[width=\columnwidth]{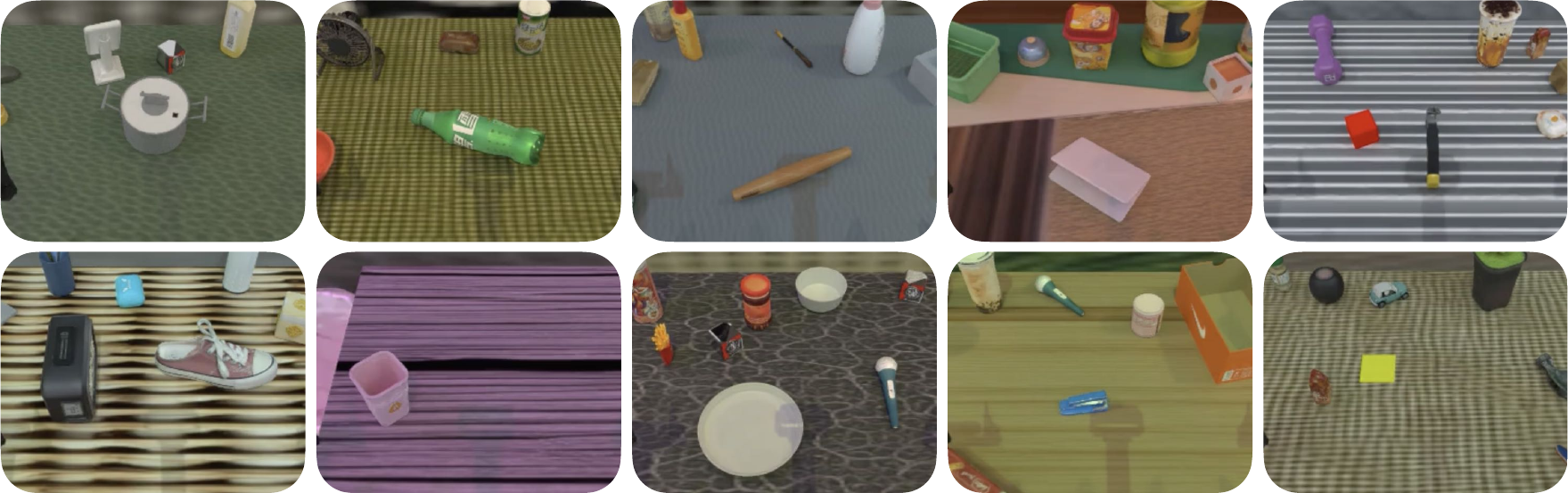}
    \caption{\textbf{Simulation Tasks with Domain Randomization.} The following tasks are used for policy evaluation. Top: \texttt{Lift Pot}, \texttt{Adjust Bottle}, \texttt{Grab Roller}, \texttt{Open Laptop}, and \texttt{Beat Block Hammer}; Bottom: \texttt{Click Alarmclock}, \texttt{Dump Bin Bigbin}, \texttt{Handover Mic}, \texttt{Press Stapler}, and \texttt{Stamp Seal}. See the Appendix for details on domain randomization.}
    \label{fig:sim_tasks}
\end{figure}

\section{Simulation Experiments}
\label{sec:sim_exps}

\subsection{Experiment Setup}
\label{sec:baselines}
We conduct simulation experiments on the open-source RoboTwin~\cite{chen2025robotwin} benchmark for bimanual manipulation using bimanual PiPER arms. We randomly select the following ten representative tasks from the full task suite for evaluation (see~\autoref{fig:sim_tasks} for an overview):
\begin{itemize}
\item \texttt{\textbf{Lift Pot}}: Lift a pot with both arms.
\item \texttt{\textbf{Adjust Bottle}}: Reorient a bottle to an upright pose.
\item \texttt{\textbf{Grab Roller}}: Grasp and pick up a roller.
\item \texttt{\textbf{Open Laptop}}: Open the lid of a closed laptop.
\item \texttt{\textbf{Beat Block Hammer}}: Strike a block with a hammer.
\item \texttt{\textbf{Click Alarmclock}}: Press the button on top of an alarm clock.
\item \texttt{\textbf{Dump Bin Bigbin}}: Empty a small bin in a large bin.
\item \texttt{\textbf{Handover Mic}}: Pass a microphone from one arm to the other.
\item \texttt{\textbf{Press Stapler}}: Press down on a stapler.
\item \texttt{\textbf{Stamp Seal}}: Pick up a seal and stamp it on a target.
\end{itemize}

The average number of time steps across these ten tasks is 147. We compare \method against the following baselines:
\begin{itemize}
    \item \textbf{ACT~\cite{Zhao-RSS-23}}: State-of-the-art RGB behavior cloning policy for bimanual manipulation, serving as our 2D imitation learning baseline.
    \item \textbf{DP3~\cite{ze2024dp3}}: Point-cloud diffusion policy that operates on 3D point clouds and requires calibrated RGB-D sensors, whereas \method uses only RGB cameras.
    \item \textbf{$\pi_{0.5}$~\cite{intelligence2025pi05}}: Open-source, pre-trained RGB-based (2D) VLA post-trained without explicit 3D input.
    \item \textbf{SmolVLA~\cite{shukor2025smolvlavisionlanguageactionmodelaffordable}}: Small, efficient, open-source, pre-trained RGB-based 2D VLA post-trained without explicit 3D input.
    \item \textbf{StereoVLA~\cite{deng2026stereovlaenhancingvisionlanguageactionmodels}}: Stereo VLA that uses the pre-trained, frozen FoundationStereo model~\cite{wen2025stereo} to extract 3D cost-volume features as an implicit stereo representation for policy learning.
    \item \textbf{$\pi_{0.5}$-\method}: Pre-trained $\pi_{0.5}$ fine-tuned with our stereo module, including the mid-training stage. Note that the stereo module is frozen during post-training.
    \item \textbf{$\pi_{0.5}$-\method w/o mid-training}: Pre-trained $\pi_{0.5}$ post-trained with our stereo module (without the mid-training stage). This serves as an ablation of \method. Note that the ViT and action-stereo cross-attention block are \emph{unfrozen} during post-training.
    \item \textbf{SmolVLA-\method}: Pre-trained SmolVLA fine-tuned with our stereo module, including the mid-training. Note that the stereo module is frozen during post-training.
\end{itemize}
Each method is evaluated on 25 unseen episodes with domain randomization, with results averaged over three training seeds.

\subsection{Datasets}

In mid-training, we train a pre-trained VLA with \method on 500 episodes from each of seven RoboTwin tasks in simulation with domain randomization~\cite{tobin2017domainrandomization} (see Appendix): \texttt{Click Bell}, \texttt{Handover Block}, \texttt{Move Can Pot}, \texttt{Move Playingcard Away}, \texttt{Place Bread Basket}, \texttt{Place Dual Shoes}, and \texttt{Place Fan}. The average number of time steps across these seven tasks is 175. Note that the mid-training tasks do not overlap with the evaluation tasks in terms of either tasks or objects. In post-training, we train each method on 100 clean (non-domain-randomized) episodes.

\subsection{Simulation Results}
Table~\ref{tab:sim-results} reports the test success rates of different methods in simulation. VLAs fine-tuned with \method (columns highlighted in \textcolor{figgreen}{green}) demonstrate significant performance gains, with improvements of up to 50.6\% for $\pi_{0.5}$ and 65.3\% for SmolVLA on \texttt{Handover Mic}. On average, \method improves the success rates of $\pi_{0.5}$ and SmolVLA by 25.2\% and 23.3\%, respectively. Moreover, VLAs with \method outperform the other baselines across all tasks. Qualitatively, VLAs fine-tuned with \method are more precise when grasping and manipulating objects, including reorienting, handing over, and pressing them.

With respect to $\pi_{0.5}$-\method (w/o MT), it outperforms the post-trained base $\pi_{0.5}$ on all tasks. This demonstrates that post-training alone with \method can consistently improve performance over the base VLA across a diverse set of tasks. With the inclusion of the mid-training stage, performance further improves on the majority of tasks. We attribute the performance degradation with mid-training on certain tasks to the choice of tasks used during mid-training. Specifically, we hypothesize that including mid-training tasks that are more similar to the post-training tasks in terms of task type and environmental variations may help address this issue.

\begin{table}[h!]
  \setlength\tabcolsep{3.0pt}
  \centering
    \footnotesize
    \begin{tabular}{lcc}
    \toprule
        Method & \makecell{Lift\\Pot} & \makecell{Open\\Laptop} \\ 
    \midrule
    VLM with Implicit Stereo Features                 & 32.0 & 20.0 \\ 
    \rowcolor{lightgreen} ASCA with Explicit Stereo Features (Ours) & \textbf{66.7} & \textbf{72.0} \\ 

    \midrule

    Mid-training without MAE & 57.3 & 90.7 \\
    \rowcolor{lightgreen} Mid-training with MAE (Ours) & \textbf{61.3} & \textbf{92.0} \\
    \bottomrule
    \end{tabular}
  \caption{
    \textbf{Ablation Studies of \method.} We compare the success rates (\%) on two RoboTwin tasks between the VLM approach with implicit stereo features and our proposed action-stereo cross-attention (ASCA) approach with explicit stereo features (rows 1 and 2), as well as between mid-training without and with the masked autoencoding (MAE) objective (rows 3 and 4). All experiments use the pre-trained $\pi_{0.5}$ as the base policy. Results are averaged over three training seeds.
  }
  \label{tab:ablation}
  \vspace{-10pt}
\end{table}

\begin{figure*}[t]
    \centering
\includegraphics[width=1.0\textwidth]{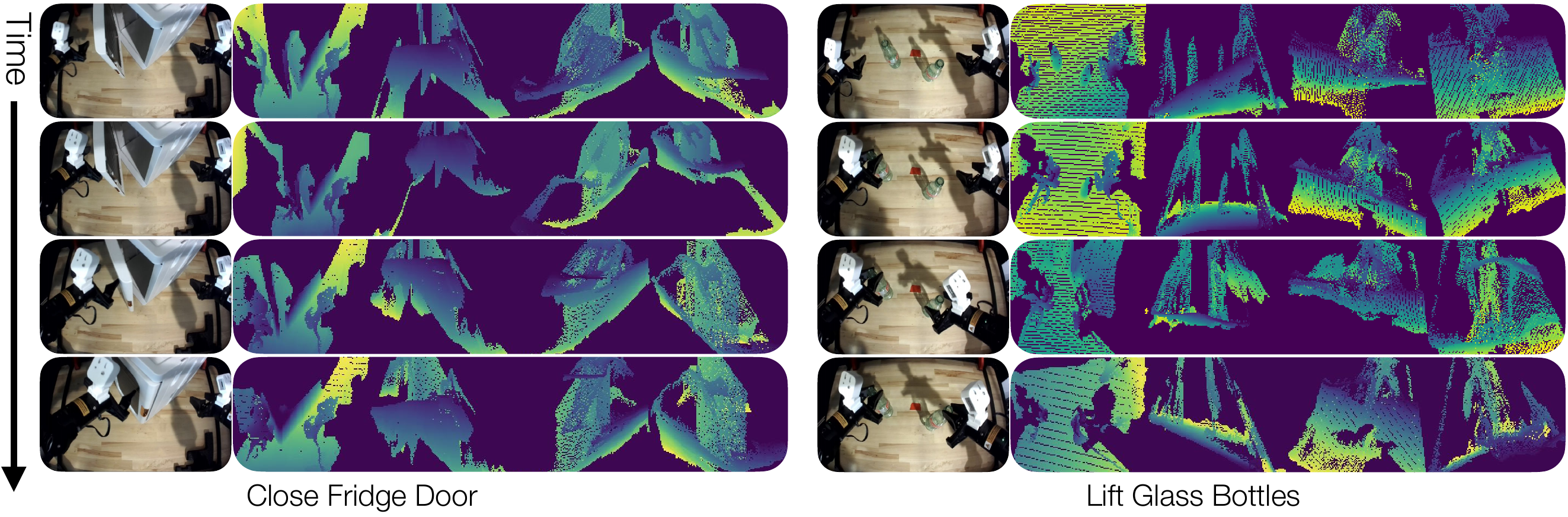}
    \caption{\textbf{Real-World Rollouts with Multi-View Observations.} Examples of successful $\pi_{0.5}$-\method rollouts on two real-world tasks using the bimanual AgileX PiPER platform. Each column shows a trajectory over time, with each row containing the left stereo image and its corresponding multi-view observations (Top, Front, Left, Right) at a given timestep. The left column shows \texttt{\textbf{Close Fridge Door}}, in which the left arm closes the fridge door; the right column shows \texttt{\textbf{Lift Glass Bottles}}, in which both arms grasp and lift two glass bottles.
    }
    \label{fig:multiview_trajectories}
\end{figure*}

\subsection{Ablation Studies}

\textbf{VLM with Implicit Stereo Features vs. Action-Stereo Cross-Attention with Explicit Stereo Features.} Prior works, namely StereoVLA~\cite{deng2026stereovlaenhancingvisionlanguageactionmodels} and StereoPolicy~\cite{han2026stereopolicyimprovingroboticmanipulation}, both concatenate implicit stereo tokens with image and language tokens as inputs to the VLM for policy learning. StereoVLA obtains these implicit features from the 3D cost-volume features of a pre-trained stereo foundation model, whereas StereoPolicy obtains them from its Stereo Feature Extractor. In this work, we demonstrate how to construct explicit stereo tokens and leverage them directly in the action expert for policy learning. As shown in Table~\ref{tab:ablation}, our approach outperforms the VLM approach with implicit stereo features when used during post-training, without retraining the VLA or requiring any additional training prior to post-training. Our approach is easier for the policy to optimize because the action expert is substantially smaller than the VLM, while the explicit multi-view observations reduce the overhead of inferring geometry from stereo image pairs or 3D cost volumes.

\textbf{Mid-training with and without MAE.} We ablate \method during mid-training with and without the additional masked autoencoding (MAE) objective, which enables the policy to learn from incomplete observations while simultaneously reconstructing the masked observations. This is especially useful for our single-stereo-camera setup, where the reconstructed point cloud often contains holes due to occlusions. This training objective forces the policy to generate robot actions conditioned on imperfect observations. As shown in Table~\ref{tab:ablation}, mid-training with MAE consistently outperforms the variant without MAE.

\section{Real-World Experiments}

\subsection{Real-World Experiment Setup}

We use two AgileX PiPER robot arms and a ZED Mini 2 camera on a tabletop platform, as shown in the top-left ``Stereo Camera'' image in~\autoref{fig:method}.
We perform experiments using the following tasks: 
\begin{itemize}
\item \texttt{\textbf{Stack Two Bowls}}: Stack two bowls on top of each other in the correct order.
\item \texttt{\textbf{Lift Glass Bottles}}: Pick up two transparent glass bottles from the table.
\item \texttt{\textbf{Close Fridge Door}}: Close the fridge door using the left arm.
\end{itemize}
We select \texttt{Stack Two Bowls} to demonstrate that our method works on standard bimanual manipulation tasks, \texttt{Lift Glass Bottles} to evaluate stereo-based depth perception on transparent objects, and \texttt{Close Fridge Door} to show that our method can also handle articulated objects.

\begin{table}[t]
  \setlength\tabcolsep{3.0pt}
  \centering
    \footnotesize
    \begin{tabular}{lccc}
    \toprule
        Method & \makecell{Stack Two\\Bowls} & \makecell{Lift Glass\\Bottles} & \makecell{Close Fridge\\Door} \\ 
    \midrule
    $\pi_{0.5}$ (Baseline)          & 2 / 20  & 5 / 20  & 9 / 20  \\ 
    StereoVLA                       & 0 / 20  & 4 / 20      & 7 / 20      \\
    $\pi_{0.5}$-\method\ w/o MT     & 9 / 20  & 11 / 20 & 16 / 20 \\ 
    \rowcolor{lightgreen} $\pi_{0.5}$-\method\ & \textbf{13 / 20} & \textbf{18 / 20} & \textbf{19 / 20} \\ 
    \bottomrule
    \end{tabular}
  \caption{
    \textbf{Real-world Experiments.} Results comparing \method with baselines, with 20 trials per method and task combination. `w/o MT'' denotes \method without the mid-training stage.
  }
  \label{tab:real-results}
\end{table}

\subsection{Datasets}
To perform mid-training, \method requires large-scale real-world stereo bimanual datasets. Existing datasets, such as DROID~\cite{khazatsky2024droid} and RoboCOIN~\cite{RoboCOINReport}, are predominantly either single-arm or monocular. In addition, few AgileX PiPER datasets are publicly available, and those that do exist are predominantly monocular. We therefore collect as many publicly available AgileX PiPER datasets from the Internet as we can find, yielding \num{20} datasets from \num{11} contributors and a total of \num{3469} clips. We also include datasets collected using the AgileX Cobot Magic robot, which extends the AgileX PiPER with a mobile base. We provide a complete list of the datasets used in the Appendix.

After collecting these datasets, we feed the monocular left views into StereoEngine~\cite{chen2027stereoengine} to generate a large-scale stereo dataset consisting of left-right image pairs while preserving the action labels from the original data. We then use the resulting dataset for policy mid-training.

For post-training, an experienced roboticist used a Meta Quest 2 to collect teleoperation data. Because each task has its own difficulty, we collect a different number of demonstrations for each task: \num{200} for \texttt{Stack Two Bowls}, \num{50} for \texttt{Lift Glass Bottles}, and \num{20} for \texttt{Close Fridge Door}. 

\subsection{Real-World Results}
Real-world results are presented in Table~\ref{tab:real-results}, with example rollouts of $\pi_{0.5}$-\method shown in \autoref{fig:multiview_trajectories}. \method demonstrates substantially improved robustness across all tasks compared to baselines.
Aggregated over the three tasks, $\pi_{0.5}$-ExStereo succeeds on 50/60 trials (83.3\%) versus
16/60 (26.7\%) for the $\pi_{0.5}$ baseline, with $\pi_{0.5}$-\method\ w/o MT in between at 36/60
(60.0\%). StereoVLA, which uses implicit stereo features, performs worst at 11/60 (18.3\%),
falling below the RGB-only $\pi_{0.5}$ baseline on every task.

We hypothesize that StereoVLA has weaker generalization capabilities than $\pi_{0.5}$, which is pre-trained on a larger and more diverse dataset. Combined with its VLM-based approach using implicit stereo features, post-training on a limited number of task demonstrations may be insufficient for StereoVLA to reach its optimal performance.
The largest gain is on \texttt{Lift Glass Bottles}, where \method reaches 18/20 while the $\pi_{0.5}$
baseline reaches only 5/20 and StereoVLA only 4/20.

\texttt{\textbf{Stack Two Bowls:}} The baseline methods frequently failed to grasp the bowls, either closing the gripper beside or above them. $\pi_{0.5}$-\method\ w/o MT and StereoVLA reliably grasped the left bowl but often failed to grasp the right bowl and stack it on top. In contrast, the RGB-only $\pi_{0.5}$ baselines failed to grasp either bowl, highlighting the importance of 3D perception for successfully grasping these objects.

\begin{figure}[t]
    \centering
    \includegraphics[width=0.5\textwidth]{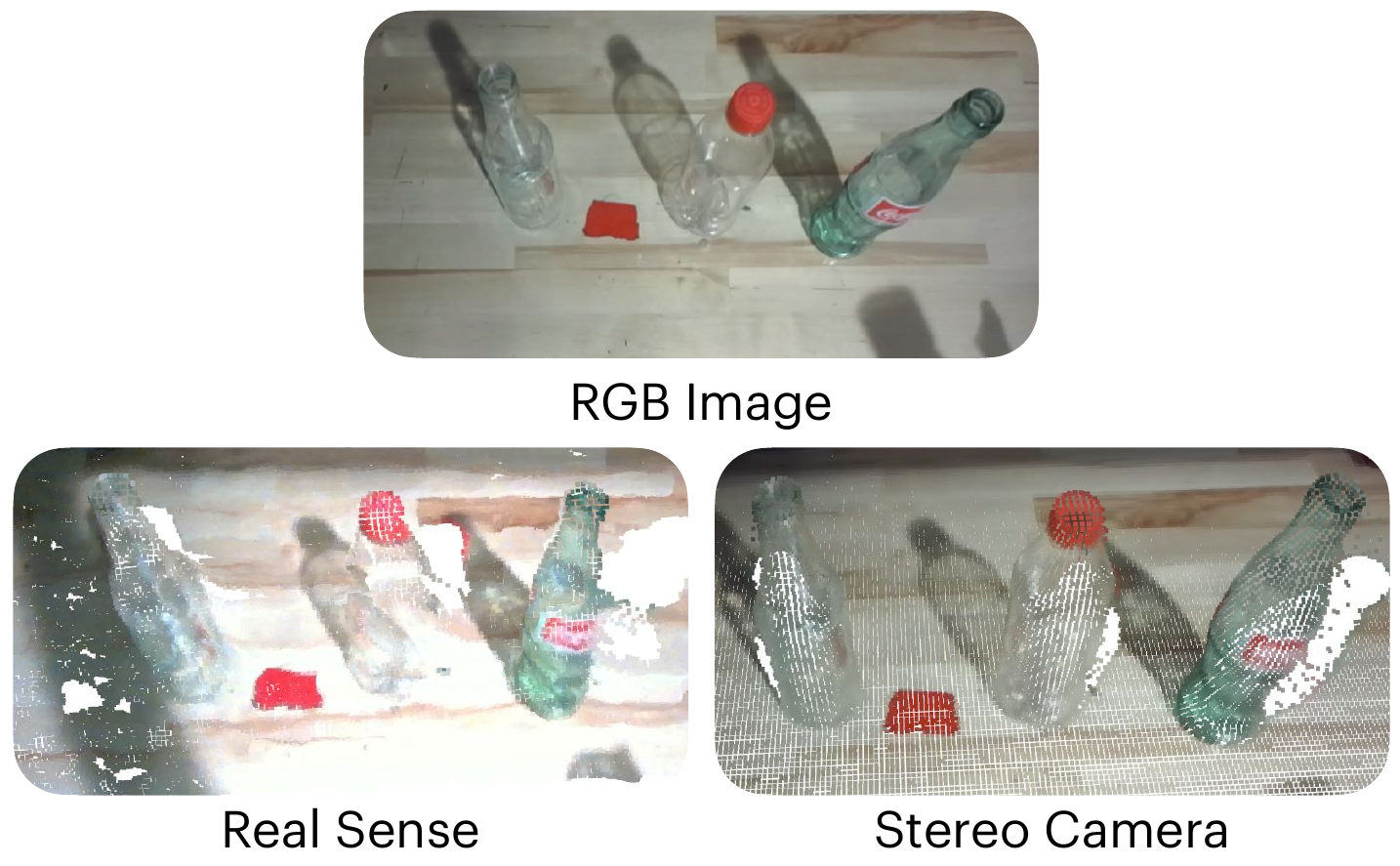}
        \caption{\textbf{Point Cloud Reconstruction of Transparent Objects.} The bottom-left image shows the point cloud captured by a RealSense RGB-D camera, which contains substantial noise and missing geometry due to the transparent objects. In contrast, the bottom-right image shows the point cloud reconstructed from a stereo image pair captured by a ZED Mini using Fast-FoundationStereo~\cite{wen2026fastfoundationstereo}, yielding cleaner and more geometrically consistent 3D points.}
      \vspace*{-10pt}
    \label{fig:transparent_comparison}
\end{figure}

\texttt{\textbf{Lift Glass Bottles:}} 
The most common failure modes for the baselines were failing to pick up the bottles and knocking them over with the gripper. In contrast, $\pi_{0.5}$-\method is the most reliable policy for precisely grasping and lifting bottles. Because this task requires precise gripper positioning, incorporating 3D spatial information about the objects is essential, as demonstrated by the improved performance of $\pi_{0.5}$ with our 3D perception augmentation.
Refer to \autoref{fig:transparent_comparison} for point cloud reconstructions of transparent objects using Fast-FoundationStereo~\cite{wen2026fastfoundationstereo}.

\texttt{\textbf{Close Fridge Door:}} The baseline methods often misjudged the distance to the door and missed it entirely, driving the arms straight into the fridge. The RGB-only $\pi_{0.5}$ baseline, in particular, tended to miss the door by reaching too far to the right or too high. In some cases, the arm applied excessive force and tipped the fridge over, which we counted as a failure. With accurate 3D perception, \method contacts the door at the correct location and closes it consistently.

\section{Limitations} 
First, \method depends on a pre-trained stereo foundation model~\cite{wen2026fastfoundationstereo}, which may degrade on distant, textureless, and thin structures, and these errors may affect the reconstructed multi-view observations. Second, the projected 3D point cloud is derived from a single stereo viewpoint, so occluded regions leave holes across all four rendered views. Third, our multi-view rendering assumes a planar surface, such as a tabletop, to define the world frame, which may not hold in cluttered or non-tabletop scenes. Finally, our mid-training data comes from a single benchmark and a single bimanual embodiment. Scaling to larger stereo robot datasets and more diverse embodiments remains an important direction for future work.

\section{Conclusion}
We present \method, a stereo module that equips any pre-trained 2D VLA with 3D perception using only a stereo camera by reconstructing explicit scene geometry from a stereo pair with an off-the-shelf stereo foundation model, rendering the geometry into a set of multi-view observations, and injecting the resulting stereo tokens into the base policy through our proposed action-stereo cross-attention mechanism. We mid-train the VLA with \method on large-scale stereo data using a self-supervised objective and further post-train the model on a small set of task demonstrations while keeping \method frozen. Across simulation and real-world experiments, VLAs fine-tuned with \method substantially outperform their base policies, as well as RGB-, point-cloud-, and stereo-based baselines. Furthermore, we demonstrate that explicitly representing stereo geometry and leveraging it in the action expert is more effective than the standard VLM approach with implicit stereo geometry for direct fine-tuning. These results demonstrate that incorporating stereo vision is a promising approach for augmenting existing 2D VLAs with 3D spatial perception.

\section{Acknowledgment}

Daniel Seita acknowledges generous support from Samsung Research America, NSF grant \#2447397, and the NVIDIA Academic Grant Program Award.

\bibliographystyle{IEEEtran}
\bibliography{main}

\clearpage
\appendix

\subsection{Implementation Details}

\noindent\textbf{\method.} The \method module adds $21.2$M parameters to a VLA ($3.2$M for the multi-view ViT, $17.3$M for the action-stereo cross-attention projections, and $0.6$M for the MAE decoder, which is discarded after mid-training). This is a minimal addition: for reference, $\pi_{0.5}$ has a $2$B-parameter VLM backbone and a $300$M-parameter action expert. We mid-train the VLA with \method for $300,000$ training steps, followed by $30,000$ steps of post-training. For $\pi_{0.5}$ and SmolVLA, we use their default hyper-parameters: those from RoboTwin for $\pi_{0.5}$ and from the LeRobot GitHub repository for SmolVLA. We do not perform any hyper-parameter tuning and directly use the checkpoint from the final training step for evaluation. See Table~\ref{tab:exstereo_hyperparams} for the hyper-parameters used in the \method module.

\noindent\textbf{StereoVLA, ACT, and DP3.} As with \method, we do not perform any hyperparameter tuning and use the default hyper-parameters from their respective codebases: the official StereoVLA GitHub repository for StereoVLA and RoboTwin for ACT and DP3. StereoVLA is trained for $30,000$ steps, while ACT and DP3 are trained for $6,000$ and $3,000$ epochs, respectively.

\begin{table}[H]
\centering
\footnotesize
\caption{\method{} module, shared by
$\pi_{0.5}$-\method{} and SmolVLA-\method{}.}
\label{tab:exstereo_hyperparams}
\begin{tabularx}{\columnwidth}{@{}lY@{}}
\toprule
\textbf{Hyper-parameter} & \textbf{Value} \\
\midrule
\multicolumn{2}{@{}l}{\textit{Point Cloud Reconstruction}} \\
Voxel downsampling     & $0.001$\,m \\
Radius outlier removal & 30 points within $0.03$\,m \\
Table-plane RANSAC     & threshold $0.01$\,m, $n = 3$, 1000 iterations \\
Canonical frame        & table plane horizontal, surface at $z = 0$ \\
\addlinespace[2pt]
\multicolumn{2}{@{}l}{\textit{Orthographic Rendering}} \\
Camera distance        & $0.6$\,m lateral, $0.5$\,m top \\
Tile resolution        & $128\times128$ per view \\
Channels               & $[\,\text{depth},\,x,\,y,\,z,\,\text{valid}\,]$ \\
Background fill        & $-1.0$ (channels 1--4), $0.0$ (valid mask) \\
Output tensor          & $128\times512\times5$ \\
\addlinespace[2pt]
\multicolumn{2}{@{}l}{\textit{Multi-View ViT Encoder}} \\
Patch size             & $16\times16$ \\
Number of patches      & 256 ($8\times32$) \\
Embedding width        & 256 \\
Depth                  & 4 blocks \\
Attention heads        & 4 \\
Positional embedding   & learned \\
\addlinespace[2pt]
\multicolumn{2}{@{}l}{\textit{MAE objective (mid-training only)}} \\
Mask ratio             & $0.30$ ($\lfloor 0.3 \cdot 256 \rfloor = 76$ patches) \\
Loss weight $\lambda_{\mathrm{MAE}}$ & $0.01$ \\
Decoder depth / width / heads & 2 / 128 / 4 \\
\bottomrule
\end{tabularx}
\end{table}

\subsection{Domain Randomization}

We use the default domain randomization settings provided in the RoboTwin documentation. Specifically, we enable random backgrounds by applying random textures to the table and background, and add table clutter using distractor objects to simulate cluttered environments. The clean-background rate is set to $0.02$, such that $2\%$ of samples retain a clean background despite background randomization. We randomly perturb the table height by up to $\pm 0.03$\,m while keeping the head-camera position fixed with no random displacement. We also enable randomized lighting, with a crazy-light rate of $0.02$, corresponding to a $2\%$ probability of applying extreme lighting conditions.

\subsection{Additional Real-World Implementation Details}
\label{additionaldetails}
\begin{figure}[t]
    \centering
    \includegraphics[width=0.5\textwidth]{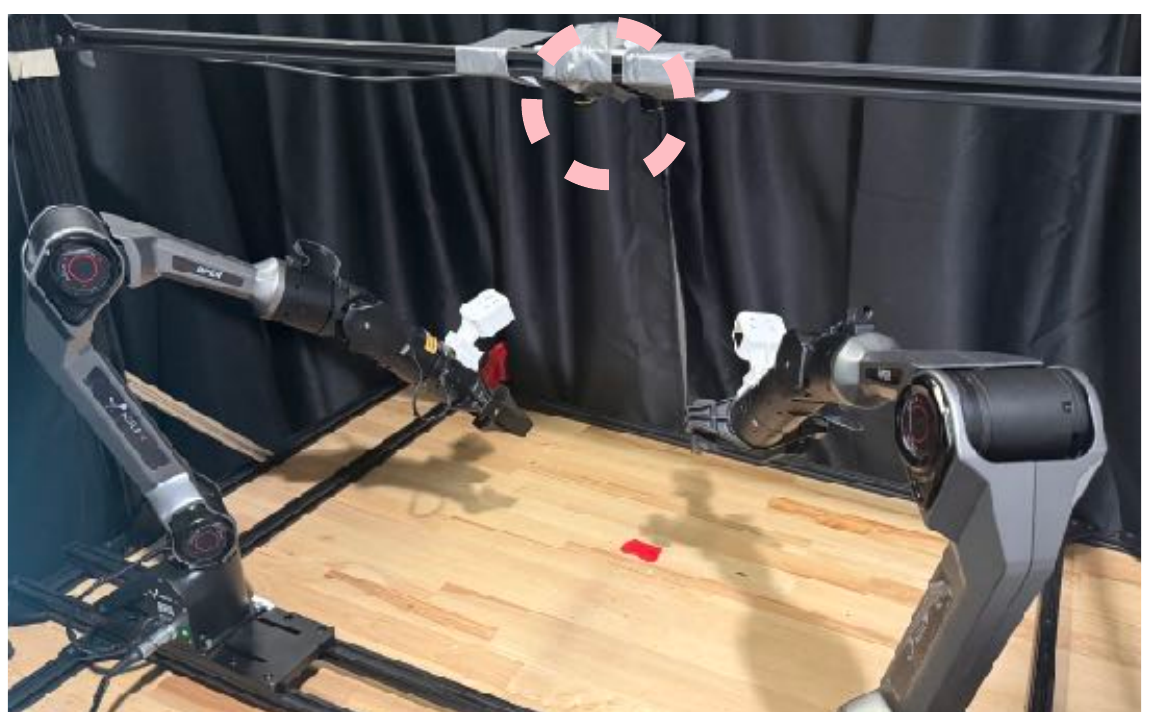}
    \caption{\textbf{Real-World Setup.} Our real-world platform with two AgileX PiPER arms. The pink dashed circle marks the ZED Mini 2 stereo camera used to capture the stereo pairs.}
      \vspace*{-10pt}
    \label{fig:realworldsetup}
\end{figure}

For the real-world setup, we use a bimanual AgileX PiPeR platform, as shown in Figure~\ref{fig:realworldsetup}. For the \texttt{Stack Two Bowls} task, we use 3D-printed bowls. For the \texttt{Lift Glass Bottles} task, we use commercially available beverage bottles. For the \texttt{Close Fridge Door} task, we use a toy refrigerator door.

For the mid-training stage of \method, we provide a list of the monocular datasets used in our experiments. These datasets were converted into stereo pairs using StereoEngine~\cite{chen2027stereoengine}, with links to the original data sources provided here:
\begin{itemize}
\item \href{https://huggingface.co/datasets/Autobrik/bi_piper_pick_and_place_fruits_mantra}{Autobrik/bi\_piper\_pick\_and\_place\_fruits\_mantra}
\item \href{https://huggingface.co/datasets/JohnnyXing/001_2026041415_bipiper}{JohnnyXing/001\_2026041415\_bipiper}
\item \href{https://huggingface.co/datasets/csjmacmi/piperx_pick_and_place_v1_4}{csjmacmi/piperx\_pick\_and\_place\_v1\_4}
\item \href{https://huggingface.co/datasets/gauravpradeep/t01_piper_pick_and_place_bimanual_lerobot}{gauravpradeep/t01\_piper\_pick\_and\_place\_bimanual\_lerobot}
\item \href{https://huggingface.co/datasets/mjung11/nc_bi_piper_pot_car_apache_merged}{mjung11/nc\_bi\_piper\_pot\_car\_apache\_merged}
\item \href{https://huggingface.co/datasets/moonyh1230/piper_wooden_block_dual_realsense_0922}{moonyh1230/piper\_wooden\_block\_dual\_realsense\_0922}
\item \href{https://huggingface.co/datasets/RoboSynChallenge/Piper_real_pour_dual}{RoboSynChallenge/Piper\_real\_pour\_dual}
\item \href{https://huggingface.co/datasets/RDLwicked/pick_up_green_block_drawer_piperX}{RDLwicked/pick\_up\_green\_block\_drawer\_piperX}
\item \href{https://huggingface.co/datasets/mjung11/nc_bi_piper_pp_grape_pot_depth_mj_A}{mjung11/nc\_bi\_piper\_pp\_grape\_pot\_depth\_mj\_A}
\item \href{https://huggingface.co/datasets/mjung11/nc_bi_piper_folding_mj_A}{mjung11/nc\_bi\_piper\_folding\_mj\_A}
\item \href{https://huggingface.co/datasets/csjmacmi/piperx_pick_and_place_v1}{csjmacmi/piperx\_pick\_and\_place\_v1}
\item \href{https://huggingface.co/datasets/csjmacmi/piperx_pick_and_place_v1_3}{csjmacmi/piperx\_pick\_and\_place\_v1\_3}
\item \href{https://huggingface.co/datasets/csjmacmi/eval_piperx_pick_and_place_v1_5}{csjmacmi/eval\_piperx\_pick\_and\_place\_v1\_5}
\item \href{https://huggingface.co/datasets/mjung11/bi_piper_open_tissue_test}{mjung11/bi\_piper\_open\_tissue\_test}
\item \href{https://huggingface.co/datasets/Sohaib03/piperx_pick_and_place_v1_2}{Sohaib03/piperx\_pick\_and\_place\_v1\_2}
\item \href{https://huggingface.co/datasets/12e21/bi_piper_subset}{12e21/bi\_piper\_subset}
\item \href{https://huggingface.co/datasets/richardshkim/bi_piper_dataset}{richardshkim/bi\_piper\_dataset}
\item \href{https://huggingface.co/datasets/mjung11/eval_bi_piper_tissue_xvla}{mjung11/eval\_bi\_piper\_tissue\_xvla}
\item \href{https://huggingface.co/datasets/csjmacmi/piperx_pick_and_place_v1_2}{csjmacmi/piperx\_pick\_and\_place\_v1\_2}
\item \href{https://huggingface.co/datasets/csjmacmi/eval_piperx_pick_and_place_v1_4}{csjmacmi/eval\_piperx\_pick\_and\_place\_v1\_4}
\end{itemize}

\end{document}